\documentclass{SCIS2025}
\usepackage{amsmath}
\usepackage{amssymb}
\usepackage{mathtools}
\usepackage{amsthm}
\usepackage{cite}
\usepackage{xcolor}
\usepackage{sverb, longtable}
\usepackage{rotating}
\usepackage{subcaption}
\usepackage{cleveref}

\usepackage{epstopdf}
\usepackage{url}
\usepackage{multirow}%
\usepackage{amssymb}%
\usepackage{amsthm}%
\usepackage{mathrsfs}%
\usepackage{amsmath}
\usepackage{mathtools}
\usepackage{booktabs}%
\usepackage{bm}
\usepackage{dsfont}
\usepackage{float}

\usepackage{pgfplots}
\usetikzlibrary{calc}
\usetikzlibrary{arrows,shapes,chains}
\usetikzlibrary{decorations.pathmorphing}
\usetikzlibrary{patterns}

\newcommand{\cb}[1]{\ifmmode {\boldsymbol{#1}}\else ${\boldsymbol{#1}}$\fi}
\newcommand{\cp}[1]{\ifmmode {\mathcal{#1}}\else ${\mathcal{#1}}$\fi}

\newcommand{\argmax}{\mathop{\mathrm{argmax}}\limits}
\begin{document}
%\oa
%%%%%%%%%%%%%%%%%%%%%%%%%%%%%%%%%%%%%%%%%%%%%%%%%%%%%%%
%%% Authors do not modify the information below
%%% 作者不需要修改此处信息
% \ArticleType{RESEARCH PAPER}
% %\SpecialTopic{}
% \Year{2025}
% \Month{January}
% \Vol{68}
% \No{1}
% \DOI{}
% \ArtNo{}
% \ReceiveDate{}
% \ReviseDate{}
% \AcceptDate{}
% \OnlineDate{}
% \AuthorMark{}
% \AuthorCitation{}
%%%%%%%%%%%%%%%%%%%%%%%%%%%%%%%%%%%%%%%%%%%%%%%%%%%%%%%

%%% title: 标题
%%%   \title{title}{title for citation}
\title{Dual Teacher-Student Learning for Semi-supervised Medical Image Segmentation}{Dual Teacher-Student Learning for Semi-supervised Medical Image Segmentation}

%%% Corresponding author: 通信作者
%%%   \author[number]{Full name}{{email@xxx.com}}
%%% General author: 一般作者
%%%   \author[number]{Full name}{}
%%% Equal Contribution: 同等贡献作者
%%%   \author[number\dag]{Full name}{}
\author[1\dag]{Pengchen~Zhang}{}
\author[1\dag*]{Alan~J.X.~Guo}{{jiaxiang.guo@tju.edu.cn}}
\author[2]{Sipin~Luo}{}
\author[3]{Zhe~Han}{}
\author[2]{Lin~Guo}{}

%%% Authors' contribution. 同等贡献声明
\contributions{These authors contributed equally to this work.}

%%% Address. 地址
%%%   \address[number]{Affiliation, City Postcode, Country}
\address[1]{Center for Applied Mathematics, Tianjin University, Tianjin 300072, China}
\address[2]{Department of Radiology, Tianjin Hospital of Tianjin University, Tianjin 300211, China}
\address[3]{Department of Hip Trauma, Tianjin Hospital of Tianjin University, Tianjin 300211, China}

%%% Abstract. 摘要
\abstract{
Semi-supervised learning reduces the costly manual annotation burden in medical image segmentation. 
A popular approach is the mean teacher (MT) strategy, which applies consistency regularization 
using a temporally averaged teacher model. 
In this work, the MT strategy is reinterpreted as a form of self-paced learning in the context of supervised learning, 
where agreement between the teacher's predictions and the ground truth implicitly guides 
the model from easy to hard. 
Extending this insight to semi-supervised learning, 
we propose dual teacher-student learning (DTSL). 
It regulates the learning pace on unlabeled data using two signals: 
a temporally averaged signal from an in-group teacher and a cross-architectural signal from a student in a second, 
distinct model group. 
Specifically, a novel consensus label generator (CLG) creates the pseudo-labels from the agreement between these two signals, 
establishing an effective learning curriculum. 
Extensive experiments on four benchmark datasets demonstrate that the proposed method 
consistently outperforms existing state-of-the-art approaches. 
Remarkably, on three of the four datasets, 
our semi-supervised method with limited labeled data surpasses its fully supervised counterparts, 
validating the effectiveness of our self-paced learning design. 
}

%%% Keywords. 关键词

\maketitle

%%%%%%%%%%%%%%%%%%%%%%%%%%%%%%%%%%%%%%%%%%%%%%%%%%%%%%%
%%% The main text. 正文部分
%%%%%%%%%%%%%%%%%%%%%%%%%%%%%%%%%%%%%%%%%%%%%%%%%%%%%%%
\section{Introduction}\label{sec:introduction}
Medical image segmentation is a vital component of computer-aided 
diagnosis~\cite{ronneberger2015u,cao2022swin,valanarasu2022unext,dong2025shape}. 
A key challenge in this domain is the need for precise annotations, 
which must be performed by experienced experts. 
This manual annotation process is both laborious and time-consuming, 
creating a significant bottleneck for developing effective methods~\cite{tajbakhsh2020embracing}. 
To address this limitation, semi-supervised medical image segmentation (SSMIS) 
has emerged as a promising and efficient approach. 
It leverages large volumes of unlabeled data to augment a limited set of labeled samples, 
reducing the reliance on exhaustive manual annotation~\cite{li2020shape,wu2022cross,zhang2023multi,he2024pair}. 

Recently, managing the learning pace~\cite{kumar2010self} of model training 
has garnered attention in semi-supervised learning,
whether through explicit or implicit mechanisms. 
For instance, in the context of open-world semi-supervised learning, methods have been introduced to explicitly balance learning between 
seen and unseen classes. 
These techniques include using an adaptive threshold with distribution alignment~\cite{guo2022robust} 
or employing an adaptive synchronizing margin loss and confidence-based clustering~\cite{ye2024bridging}. 

In the context of SSMIS, 
techniques implicitly incorporate the principles of curriculum learning~\cite{bengio2009curriculum}. 
For instance, using a confidence threshold on weakly perturbed samples to filter noisy pseudo-labels~\cite{yang2023revisiting}
inherently controls the learning pace, progressing from high-confidence to more challenging regions. 
Similarly, the approach proposed in \cite{chi2024adaptive} leverages confidence to generate samples 
tailored to different training stages by including or excluding low-confidence areas, 
which implicitly establishes a confidence-based curriculum. 

The mean teacher (MT) strategy, widely recognized for enforcing consistency regularization~\cite{tarvainen2017mean, su2019local, li2021hierarchical, tan2022hyperspherical, fan2023revisiting, liu2022perturbed}, 
also implicitly incorporates a curriculum learning mechanism to some extent, as will be analyzed later. 
This underlying principle also extends to related methods like dual teacher~\cite{liu2022perturbed} and dual student~\cite{na2023switching}, 
although it is rarely acknowledged in their design. 

\begin{figure*}[!htb]    
  \centering
  \includegraphics[width=0.77\linewidth]{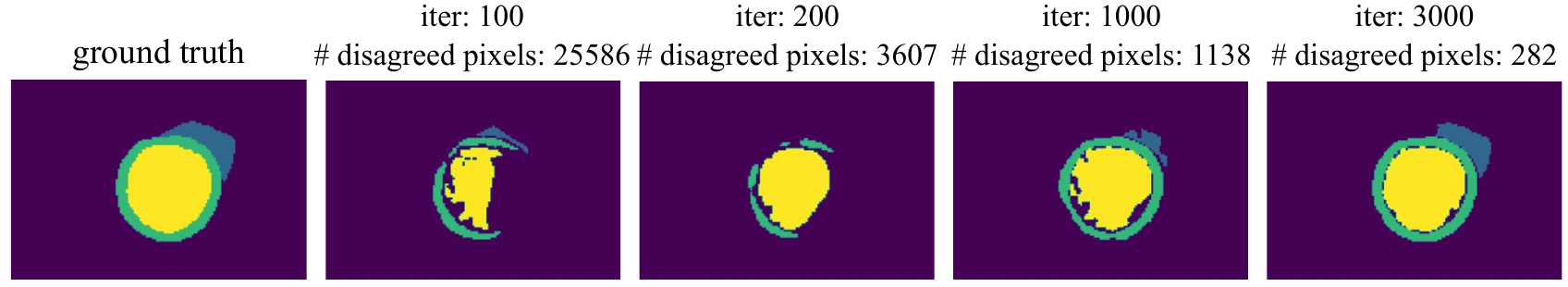}
%   \caption{The variation of consistent regions during model training in ACDC dataset(The first number represents the number of inconsistent pixel predictions between the two models, and the second number represents the number of iterations.The first line shows the consistency relationship between student model predictions and teacher model generated pseudo labels, while the second line shows the consistency relationship between student model predictions and GT).}
%   \caption{The agreement between teacher-generated labels 
%   and ground truth labels 
%   across training iterations in supervised segmentation with the MT strategy. 
%   The regions of agreement expand from the object center to the borders as training progresses. }
  \caption{Agreement between teacher-generated labels and ground truth labels across training iterations 
  in a supervised setting with the MT strategy. 
  The regions of agreement expand from the object's center to its borders as training progresses. 
  This suggests that the MT strategy may implicitly introduce a curriculum, 
  guiding the model to learn from easier to more complex regions.}
  \label{fig:process}
\end{figure*}

In the MT framework, 
two models with identical architecture are instantiated as the teacher and student. 
The student is trained using both human-annotated labels and pseudo-labels generated by the teacher.
The teacher model is updated via an exponential moving average (EMA) of the student’s weights, 
\begin{equation}\label{eqn:EMA}
    \Theta^T_t= \omega \Theta^T_{t-1}+(1-\omega)\Theta^S_t, 
\end{equation}
where $\Theta^S_t$ is the weight of the student model at step $t$, 
and $\Theta^T_{t-1}$ is the weight of the teacher model at step $t-1$. 
This strategy has demonstrated strong performance in SSMIS and 
has inspired numerous successful extensions~\cite{shi2021inconsistency,bai2023bidirectional,chi2024adaptive}. 

Applying MT strategy on the supervised learning implicitly introduces a mechanism similar to 
self-paced learning~\cite{kumar2010self} or curriculum learning~\cite{bengio2009curriculum}. 
% Taking a supervised learning task using the MT strategy as an example, 
Let $\bm{y}_{\mathrm{gt}}$ denote the ground truth segmentation label of a sample $\bm{x}$, 
$\hat{\bm{y}}$ the student output, and $\hat{\bm{y}}_{\mathrm{t}}$ the teacher-generated pseudo-label. 
The training objective for the student is 
\begin{equation}\label{loss:seperate}
    \mathcal{L}=\mathcal{L}(\hat{\bm{y}},\bm{y}_{\mathrm{gt}})+ 
    \lambda\mathcal{L}(\hat{\bm{y}},\hat{\bm{y}}_{\mathrm{t}}),
\end{equation}
which can be reformulated based on whether the teacher's prediction agrees with the ground truth, i.e., 
$\hat{\bm{y}}_{\mathrm{t}} = \bm{y}_{\mathrm{gt}}$
\begin{align}
    \mathcal{L} &= \mathds{1}({\hat{\bm{y}}_{\mathrm{t}} = \bm{y}_{\mathrm{gt}}})
    (1+\lambda) \mathcal{L}(\hat{\bm{y}},\bm{y}_{\mathrm{gt}}) \notag \\
    &+ \mathds{1}({\hat{\bm{y}}_{\mathrm{t}} \neq \bm{y}_{\mathrm{gt}}}) (\mathcal{L}(\hat{\bm{y}},\bm{y}_{\mathrm{gt}})+ 
    \lambda\mathcal{L}(\hat{\bm{y}},\hat{\bm{y}}_{\mathrm{t}})), \label{eqn:mt1}
\end{align}
where $\mathds{1}(\cdot)$ is the indicator function, applied pixel-wise, that activates the corresponding term based on whether 
$\hat{\bm{y}}_{\mathrm{t}}$ matches ground truth $\bm{y}_{\mathrm{gt}}$. 
When the teacher label (temporally averaged signal) consensus with the ground truth, 
the loss simplifies to
\begin{equation}\label{eqn:mt2}
    \mathcal{L}= (1+\lambda)\mathcal{L}(\hat{\bm{y}},\bm{y}_{\mathrm{gt}}), 
\end{equation}
indicating reinforcement of the ground truth signal. 
When the teacher label disagrees with the ground truth, the loss becomes
\begin{equation}\label{eqn:mt3}
    \mathcal{L}=\mathcal{L}(\hat{\bm{y}},\bm{y}_{\mathrm{gt}})+ 
    \lambda\mathcal{L}(\hat{\bm{y}},\hat{\bm{y}}_{\mathrm{t}}),
\end{equation}
which reflects competing gradients from two sources, balancing the optimization of $\hat{\bm{y}}$ in different directions. 

Therefore, as illustrated in 
\Cref{eqn:mt1,eqn:mt2,eqn:mt3}, 
signals from both the temporally averaged teacher and the ground truth label jointly 
control the learning pace, when the MT strategy is applied to supervised learning. 
During the early epochs of training, the teacher model tends to make confident predictions on easier regions, 
causing the optimization target \Cref{loss:seperate} to focus more on these simpler tasks. 
As training progresses, the agreement between the teacher model and the ground truth gradually extends to more difficult regions, 
thereby guiding the student model to tackle increasingly challenging examples. 

\begin{table}[!htb]
\centering
\setlength{\tabcolsep}{3.2pt}
\caption{Comparison of supervised segmentation performance with and without the MT strategy. 
The MT strategy enhances performance in a fully supervised setting, 
suggesting it introduces beneficial effects that may be interpreted as consistency regularization or implicit self-paced learning.}
\label{tab:sup}
\begin{tabular}{r|cccc|cc}
    \toprule
    & \multicolumn{4}{c|}{ACDC} & \multicolumn{2}{c}{PROMISE12} \\
    % \cmidrule{2-7}
    & DSC$\uparrow$ & Jaccard$\uparrow$ & 95HD$\downarrow$ & ASD$\downarrow$ & DSC$\uparrow$  & ASD$\downarrow$\\
    \midrule
    U-Net (w/o MT)  & 91.44 & 84.59 & 4.30 & 0.99& 84.76 &  1.58\\
    U-Net (w/ MT)  & \textbf{91.70} & \textbf{85.02} & \textbf{1.15} & \textbf{0.29} & \textbf{85.22} & \textbf{1.55} \\
    \bottomrule
\end{tabular}
\end{table}

To verify that the MT strategy introduces a self-paced learning effect, 
we conducted experiments in a fully supervised setting on the ACDC and PROMISE12 datasets. 
As shown in \Cref{tab:sup}, 
applying the MT strategy in this supervised-only context improves segmentation performance. 
Furthermore, \Cref{fig:process} visualizes the agreement between 
teacher-generated and ground truth labels across training iterations.
The regions of agreement progressively expand from the object's center to its more challenging borders as training advances.
This observation partly supports the hypothesis that the MT strategy implicitly controls the learning pace, 
guiding the model from easier to more difficult regions. 

However, when applied to unlabeled data, 
the standard MT strategy lacks this intrinsic self-paced mechanism due to the absence of a ground truth signal. 
To introduce a robust self-paced learning framework to SSMIS, 
this work utilizes two distinct signals to control the learning pace, 
the aforementioned temporally averaged teacher signal and a cross-architectural signal. 
The underlying hypothesis is that early in training, 
these two diverse signals are more likely to achieve consensus on simpler regions, 
which could be used to control the learning pace. 
Notably, this consensus-based mechanism is distinct from approaches that rely solely on prediction confidence 
to guide the learning pace. 

Specifically, a framework of dual teacher student learning (DTSL) is proposed, 
which employs two teacher-student groups with distinct architectures. 
Within each group, the teacher and student share the same architecture, 
and the teacher model maintains an EMA of its corresponding student's weights. 
For each student, both its in-group teacher (temporally averaged signal) and the cross-group student (cross-architectural signal)
regulate the pace of learning by using the consensus of these two signals. 
To implement this, 
a consensus label generator (CLG) is proposed to 
create pseudo-labels from regions where these two signals agree. 
Agreement is quantified by measuring the Jensen-Shannon (JS) divergence \cite{lin1991divergence} 
between the outputs of the two different models. 

The main contributions of this work are as follows:
\begin{itemize}
    \item The curriculum learning effect of the MT strategy is first analyzed in the supervised learning context 
    and then extended to the semi-supervised learning framework.
    \item Different from confidence-based learning pace control, 
    this work is the first to implement self-paced learning in SSMIS 
    by leveraging two distinct signals, 
    which provides greater flexibility in controlling the pace of learning and enhances model performance.
    \item In the context of SSMIS, outputs from cross-architectural models were first explicitly leveraged 
    for generating pseudo labels.
\end{itemize}

Extensive experiments across various methods and datasets demonstrate
that the proposed DTSL framework outperforms state-of-the-art (SOTA) approaches
in both supervised and semi-supervised medical image segmentation. 
Notably, applying DTSL in a semi-supervised setting outperforms standard supervised learning on three of the four datasets. 
To the best of our knowledge, this is the first such achievement. 
Ablation studies and hyperparameter optimization further validate the effectiveness of the proposed method.

\section{Related Works}
\label{gen_inst}

\subsection{Medical image segmentation}

Contemporary medical image segmentation aims to delineate regions of interest across various imaging modalities, 
such as magnetic resonance imaging (MRI) 
and computed tomography (CT)~\cite{cciccek20163d,dou2020unpaired,li2018h,wang2021automatic}. 
Convolutional Neural Networks (CNNs) have been foundational in this domain,
with architectures like UNet~\cite{ronneberger2015u}, 
VNet~\cite{abdollahi2020vnet}, ResUNet~\cite{diakogiannis2020resunet}, \emph{etc.}, 
achieving remarkable results. 
However, a major limitation of these supervised methods is their reliance on large quantities of precise, 
high-quality annotations—a process that demands substantial time and expertise~\cite{tajbakhsh2020embracing}. 

\subsection{Semi-supervised learning}
Semi-supervised learning aims to train models using a combination of labeled and unlabeled data \cite{bai2023bidirectional,he2024pair}. 
The main approaches include pseudo-labeling, consistency regularization, and entropy minimization. 
Pseudo-labeling methods \cite{yao2022enhancing,wang2022rethinking} 
first perform supervised learning on labeled data, 
then generate pseudo-labels for unlabeled data, and finally refine 
these pseudo-labels using strategies such as random propagation. 
Consistency regularization methods \cite{luo2021semi,wu2022exploring,xu2023dual} 
enforce the model to produce consistent predictions 
on unlabeled data under various perturbations. 
Entropy minimization methods \cite{hang2020local,vesal2021adapt} 
aim to minimize the output entropy of the model for unlabeled samples. 

A widely adopted strategy in semi-supervised learning is the MT framework~\cite{tarvainen2017mean}, 
which encourages the student model to produce predictions consistent with those of the teacher model. 
This approach has inspired numerous successful 
algorithms~\cite{su2019local,hang2020local,li2021hierarchical,liu2022perturbed,na2023switching}. 
Many of these MT strategy-based methods aims to generate high quality pseudo-labels. 
For example, Tri-U-MT\cite{wang2021tripled} introduces a triple uncertainty-guided MT framework to  improve prediction performance. 
% Moreover, due to the imperfect and inevitably erroneous pseudo-labels generated by the teacher network, 
% this may lead to training bias. Therefore, we aim to design a module that can minimize the errors in 
% pseudo-labels as much as possible.

Similarly, co-learning has also proven to be an effective approach in SSMIS~\cite{wang2023mcf,gao2023correlation}. 
These methods typically employ two sub-networks, often with distinct architectures, 
to generate pseudo-labels for each other. 
This mutual supervision process enables the models to iteratively learn more representative features from unlabeled data. 
For instance, CrossTeaching~\cite{luo2022semi} 
leverages two architecturally different networks to perform cross-pseudo supervision. 
However, a key challenge for these frameworks is maintaining training stability. 
Without the grounding influence of a temporally-averaged teacher model, 
the pseudo-labels generated by the concurrently training networks can be noisy, potentially leading to error accumulation. 
% to solve this problem, we propose two sets of teacher student models with different network structures to maximize the stability and diversity between the models.

\subsection{Self-paced learning}
Self-paced learning \cite{kumar2010self} is a learning methodology inspired by cognitive science, 
which cautiously and adaptively prioritizes learning from simple and reliable examples, 
and gradually transitions to more difficult ones. 
Common self-paced learning methods \cite{jiang2014self,jiang2015self} 
are capable of autonomously evaluating 
the difficulty of samples, making the learning process dynamic and optimizable. 
Self-paced learning has been successfully applied to a variety of tasks, 
such as saliency detection \cite{zhang2016co} and object tracking\cite{supancic2013self}. 

This self-paced learning principle has also been applied in semi-supervised segmentation tasks, 
guiding models to learn by selecting sample regions from simple to complex. 
Methods such as FixMatch~\cite{sohn2020fixmatch}, UniMatch~\cite{yang2023revisiting}, CWMS~\cite{chen2023confidence},ABD~\cite{chi2024adaptive}, 
PriorsMatch~\cite{chen2026priorsmatch}, 
\emph{etc} exemplify this approach. 
In these works, high-confidence predictions are retained as pseudo-labels, 
which implicitly controls the learning pace via a confidence-based curriculum. 
Although these methods have achieved good results, 
their reliance on prediction confidence as the sole proxy for sample simplicity is a limitation, 
as a model's self-reported confidence can often be miscalibrated and unreliable. 
% This paper proposes this proposition for the first time and achieves ideal results by more effectively controlling 
% step size through two methods: the time-lag model and the structural different model. 

\section{Method}
\label{headings}

\begin{figure*}[!htbp]    
    \centering
    \includegraphics[width=0.77\linewidth]{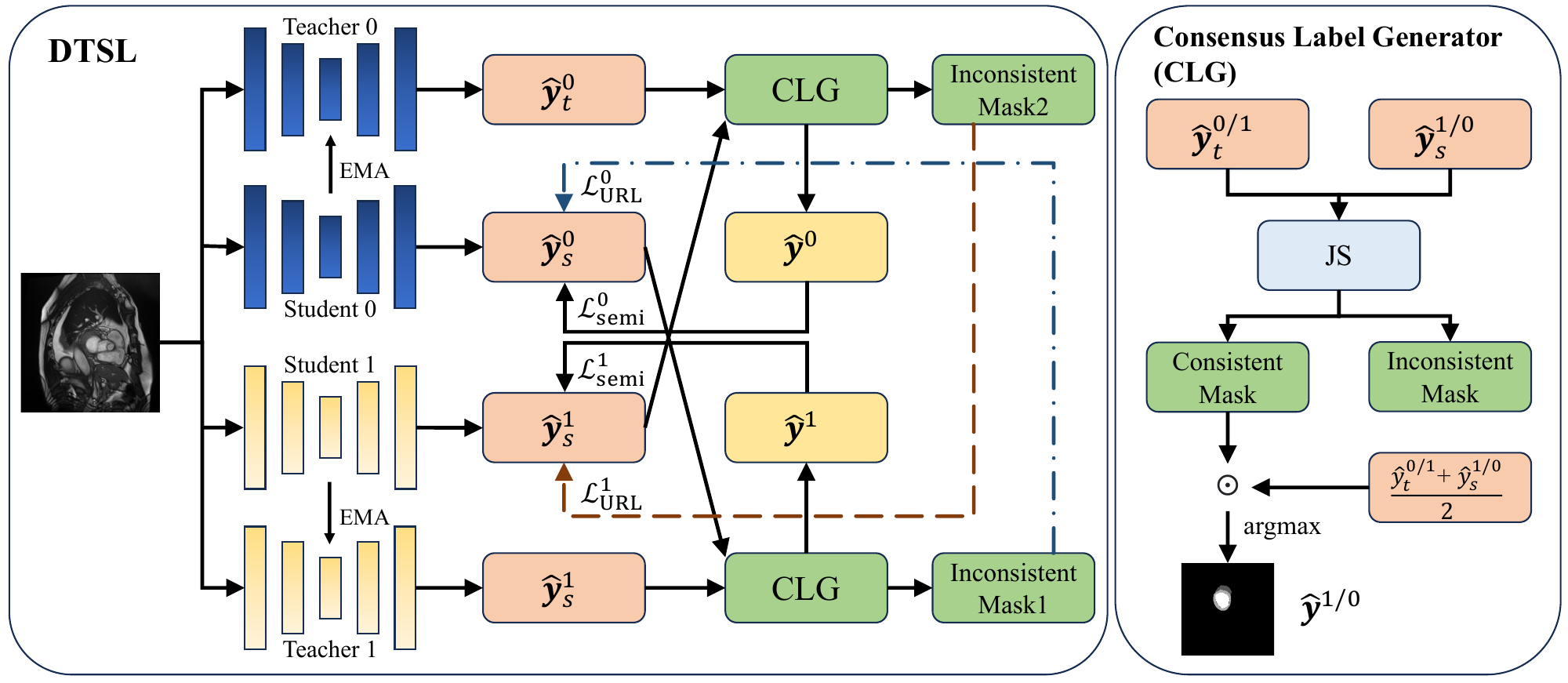}
    % \caption{Overview of the DTSL for SSMIS. Supervised part is omitted. 
    % The consensus label generator (CLG) takes predictions from teacher model and cross-grouped student model, 
    % and outputs the consistent and inconsistent parts. The consistent part is used as the pseudo label for the student model, 
    % while the inconsistent part is used by the uniform regularization. 
    % }
    \caption{Overview of the proposed DTSL framework on an unlabeled data sample. 
    The framework utilizes two teacher-student groups with distinct architectures. 
    The core component, the CLG, 
    creates a pseudo-label based on the consensus between the in-group teacher's prediction 
    and the cross-group student's prediction. 
    This pseudo-label guides the student model via a semi-supervised loss ($\mathcal{L}_{\mathrm{semi}}$). 
    For regions of disagreement, a uniform regularization ($\mathcal{L}_{\mathrm{URL}}$) is applied to handle prediction uncertainty.}
    \label{fig:dtsl_overview}
\end{figure*}

\subsection{Problem setting}
The training set $D$ for semi-supervised segmentation consists of a small labeled subset 
$D^{\ell} = \{(\bm{x}_i^{\ell}, \bm{y}_i^{\ell})\}_{i=1}^{N}$ 
and a large unlabeled subset 
$D^u=\{\bm{x}_i^u\}_{i=N+1}^{M+N}$, where $N \ll M$. 
For the labeled data, each image
$\bm{x}_i^{\ell}$ in $D^{\ell}$ has a corresponding pixel-wise annotation 
$\bm{y}_i^{\ell} \in \{0,1,\dots,K-1\}^{H\times W}$, 
where $K$ is the total number of classes and class $0$ usually denotes the background. 

The objective of semi-supervised learning is to leverage both the labeled data $D^{\ell}$ 
and the unlabeled data $D^u$ to train a model that generalizes effectively, 
making accurate predictions on the unseen test set $D^{t} = \{(\bm{x}_i^{t}, \bm{y}_i^{t})\}$.

% The objective of semi-supervised learning is to leverage both the labeled data $D^{\ell}$ 
% and the unlabeled data $D^u$ to train a model that generalizes effectively, 
% making accurate predictions on the unseen test set $D^{t} = \{(\bm{x}_i^{t}, \bm{y}_i^{t})\}$.

\subsection{Dual teacher student learning framework}
As illustrated in \Cref{fig:dtsl_overview}, 
the proposed DTSL framework employs two teacher-student groups with distinct architectures. 
Within each group, the teacher and student models share the same architecture. 

During training, the student models are updated via gradient descent on a composite loss function. 
Concurrently, the parameters of each teacher model are updated 
using the EMA strategy defined in \Cref{eqn:EMA}, 
based on their respective student's weights.

The optimization objective for a labeled sample, $\mathcal{L}^{\ell}$, is a composite loss consisting of two components, 
\begin{equation}\label{eqn:totallosssupervised}
    \mathcal{L}^{\ell}=\mathcal{L}_{\mathrm{sup}}+ \mathcal{L}_{\mathrm{pace}},
\end{equation}
where $\mathcal{L}_{\mathrm{sup}}$ is the conventional supervised loss and 
$\mathcal{L}_{\mathrm{pace}}$ is the proposed loss term to regulate the learning pace. 
The supervised component $\mathcal{L}_{\mathrm{sup}}^i$ for $\mathrm{student}_i$ from group $i, (i=0,1)$ 
is defined as a standard mixed loss of cross-entropy and Dice loss~\cite{sudre2017generalised}, 
\begin{equation}
    \mathcal{L}^i_{\mathrm{sup}} = \frac{1}{2}(
    \mathcal{L}_{\mathrm{CE}}(\hat{\bm{y}}^i_s, 
    \bm{y}_{\mathrm{gt}})+\mathcal{L}_{\mathrm{dice}}(\hat{\bm{y}}^i_s,\bm{y}_{\mathrm{gt}})), 
\end{equation}
where $\hat{\bm{y}}^i_s$ is the $i$-th student's prediction and $\bm{y}_{\mathrm{gt}}$ 
is the ground truth. 

For unlabeled samples, the optimization objective $\mathcal{L}_{\mathrm{pace}}$ consists only of the pace regulator 
\begin{equation}\label{eqn:totallossunsupervised}
    \mathcal{L}^{u}=\mathcal{L}_{\mathrm{pace}}.
\end{equation}
In this context, $\mathcal{L}_{\mathrm{pace}}$ serves a dual purpose: it controls the learning pace 
while simultaneously functioning as the mechanism for pseudo-label propagation.

\subsubsection{Consensus label generator (CLG)}
As mentioned in \Cref{sec:introduction}, the MT strategy, when applied in a fully supervised setting, 
introduces an implicit learning pace. 
This pace is regulated by the agreement between the temporally-averaged teacher model and the ground truth labels.

In this work, this self-paced concept is explicitly adapted for semi-supervised learning. 
We extend the pace-regulating mechanism to be governed by the agreement between two distinct signals, 
one from a temporally-averaged model and another from a cross-architectural model. 
This is implemented by the proposed CLG. 
% As depicted in the CLG module of \cref{fig:dtsl_overview}, 
% this component generates pseudo-labels by identifying regions of consensus between the outputs of the two models.

As illustrated in \Cref{fig:dtsl_overview}, to generate a pseudo-label for a given student, 
the inputs of CLG are sourced from its in-group teacher and the cross-group student. 
The consistency between these two predictions is used to create the pseudo-label for that student. 

Let $\bm{o}_1$ and $\bm{o}_2$ denote the outputs of the two different models, 
representing the pixel-wise predicted categorical distributions. 
The JS divergence \cite{lin1991divergence} is computed between $\bm{o}_1$ and $\bm{o}_2$ pixel-wisely, 
\begin{equation}
    \mathrm{JS}(\bm{o}_1,\bm{o}_2) = \frac{\mathrm{KL}(\bm{o}_1||\bm{m}) + \mathrm{KL}(\bm{o}_2||\bm{m})}{2}, 
\end{equation}
where $\bm{m}=\frac{\bm{o}_1+\bm{o}_2}{2}$ is the average distribution. 
JS divergence is a symmetric measure of similarity between two distributions, 
derived from the Kullback–Leibler (KL) divergence \cite{kullback1951information}, 
bounded in $[0, 1]$ when using a base-2 logarithm. 

Next, a threshold $\kappa\in [0,1]$ is applied to the resulting JS divergence map to create binary masks 
for consistent (easy) and inconsistent (difficult) regions, 
\begin{align}\label{eqn:kappa}
    \mathrm{Mask}_{\mathrm{cons}}&=\mathds{1}(\mathrm{JS}(\bm{o}_1,\bm{o}_2)<\kappa);\\
    \mathrm{Mask}_{\mathrm{diff}}&=\mathds{1}(\mathrm{JS}(\bm{o}_1,\bm{o}_2)\geq \kappa), 
\end{align}
where $\mathds{1}(\cdot)$ denotes the the pixel-wise indicator function 
that returns $1$ if the condition is true and $0$ otherwise. 

The CLG then creates the final pseudo-label. 
It uses the $\argmax$ of the average distribution of $\bm{o}_1$ and $\bm{o}_2$ 
for consistent regions (easy regions), 
and assigns the background class to all inconsistent regions (difficult regions), 
\begin{align}\label{eqn:CLG}
    &\mathrm{CLG}(\bm{o}_1,\bm{o}_2) \nonumber \\ 
    =&\argmax \left(\frac{\bm{o}_1+\bm{o}_2}{2}\odot \mathrm{Mask}_{\mathrm{cons}} 
    + \mathrm{onehot}(\bm{0}) \odot \mathrm{Mask}_{\mathrm{diff}}\right), 
\end{align}
where $\mathrm{onehot}(\bm{0})$ represents the one-hot encoding of the background class.

This approach generates pseudo-labels in a self-paced manner, 
establishing a curriculum that begins with easy regions and gradually incorporates more difficult ones as training progresses. 
Early in training, the models are more likely to achieve consensus on simple regions. 
As the models become more accurate, their agreement extends to more challenging areas, 
allowing these harder regions to be progressively incorporated into the learning process. 

\subsubsection{Pace regulator: $\mathcal{L}_{\mathrm{pace}}$}
The pace regulator $\mathcal{L}_{\mathrm{pace}}$ consists of two components, 
a conventional semi-supervised loss $\mathcal{L}_{\mathrm{semi}}$ that applies pseudo-labels to consistent (easy) regions, 
and a uniform regularization loss (URL) $\mathcal{L}_{\mathrm{URL}}$ applied to inconsistent (difficult) regions. 
The URL encourages the model to produce high-entropy (uncertain) predictions in areas where the guiding signals disagree. 

As part of $\mathcal{L}_{\mathrm{pace}}$, we define a semi-supervised loss, $\mathcal{L}_{\mathrm{semi}}$, 
based on the pseudo-labels generated by the CLG. This loss aligns with standard pseudo-labeling in semi-supervised learning frameworks. 
% Denote the predictions made by $i$-th student and $i$-th teacher as $\hat{\bm{y}}^i_s$ and $\hat{\bm{y}}^i_t$ 
% respectively, where $i=0,1$ indicates the group number. 
% As part of $\mathcal{L}_{\mathrm{pace}}$, a semi-supervised loss is defined between the student predictions and 
% the pseudo-labels generated by CLG, 
% \begin{align}
%     \mathcal{L}_{\mathrm{semi}}^0 &= 
%     \mathcal{L}_{\mathrm{dice}}(\hat{\bm{y}}^0_s, \mathrm{CLG}(\hat{\bm{y}}^1_s,\hat{\bm{y}}^0_t)), \\
%     \mathcal{L}_{\mathrm{semi}}^1 &= 
%     \mathcal{L}_{\mathrm{dice}}(\hat{\bm{y}}^1_s, \mathrm{CLG}(\hat{\bm{y}}^0_s,\hat{\bm{y}}^1_t)), 
% \end{align}
% which aligns with the common use of pseudo-labels in semi-supervised learning frameworks. 
For $\mathrm{student}_0$ from $0$-th group, the loss is computed against a pseudo-label generated from 
the consensus of its in-group teacher ($\mathrm{teacher}_0$) and the cross-group student ($\mathrm{student}_1$). 
Conversely, $\mathrm{student}_1$ is supervised by the consensus of $\mathrm{teacher}_1$ and $\mathrm{student}_0$:
\begin{align}
    \mathcal{L}_{\mathrm{semi}}^0 &= 
    \mathcal{L}_{\mathrm{dice}}(\hat{\bm{y}}^0_s, \mathrm{CLG}(\hat{\bm{y}}^1_s,\hat{\bm{y}}^0_t)), \\
    \mathcal{L}_{\mathrm{semi}}^1 &= 
    \mathcal{L}_{\mathrm{dice}}(\hat{\bm{y}}^1_s, \mathrm{CLG}(\hat{\bm{y}}^0_s,\hat{\bm{y}}^1_t)), 
\end{align}
where $\hat{\bm{y}}^i_s$ and $\hat{\bm{y}}^i_t$ denote the predictions of the student and teacher from group $i \in \{0,1\}$, 
respectively.

Motivated by \cite{song2024sdcl}, an additional loss term is applied 
to handle inconsistent regions between a student and its cross-group teacher. 
This term, referred to as the URL, 
encourages the model to reconsider its discrepant predictions by increasing the prediction's entropy, 
which is achieved by comparing the model's output distribution to a discrete uniform distribution using KL divergence.

Specifically, taking $\mathrm{student}_0$ as an example, 
the inconsistent regions is firstly identified by comparing its prediction
$\hat{\bm{y}}^0_s$ with the cross-groupe teacher's prediction $\hat{\bm{y}}^1_t$, 
\begin{equation}
    \mathrm{Mask}^0_{\mathrm{diff}}=\mathds{1}(\mathrm{JS}(\hat{\bm{y}}^0_s,\hat{\bm{y}}^1_t)\geq \kappa), 
\end{equation}
where $\mathds{1}(\cdot)$ is the pixel-wise indicator function. 
The URL for $\mathrm{student}_0$ is then computed only on these inconsistent regions, 
\begin{equation}\label{eqn:URL}
    \mathcal{L}^0_{\mathrm{URL}} = \mathrm{Mask}^0_{\mathrm{diff}} \odot \mathrm{KL}(\hat{\bm{y}}^0_s, \mathrm{Uniform}),  
\end{equation}
where $\mathrm{Uniform}$ denotes the discrete uniform distribution over all classes. 

Finally, the complete pace regulator $\mathcal{L}_{\mathrm{pace}}$ 
is defined as the weighted sum of the semi-supervised loss $\mathcal{L}_{\mathrm{semi}}$ 
and the uniform regularization loss $\mathcal{L}_{\mathrm{URL}}$, 
\begin{align}\label{eqn:pace}    
    \mathcal{L}_{\mathrm{pace}} &= \alpha \mathcal{L}_{\mathrm{semi}} + \beta \mathcal{L}_{\mathrm{URL}} \\\notag
   & = \alpha (\mathcal{L}^0_{\mathrm{semi}}+\mathcal{L}^1_{\mathrm{semi}}) + 
    \beta (\mathcal{L}^0_{\mathrm{URL}}+\mathcal{L}^1_{\mathrm{URL}}),   
\end{align}
where $\alpha$ and $\beta$ are hyperparameters that balance the contribution of the two corresponding terms. 

% It is worth noting that, ususally one hyperparameter is used to adjust the ratio of the two loss terms. 
% However, as in the supervised part in \cref{eqn:totallosssupervised}, there is three loss term. 
% So we keep two hyperparameters engaged. In the hyperparameter optimization, it could be found that the ratio of $\alpha$ 
% and $\beta$ is more crucial like one hyperparameter. 

In summary, by optimizing the composite losses defined in \Cref{eqn:totallosssupervised} and \Cref{eqn:totallossunsupervised}, 
the proposed DTSL framework is trained in a self-paced manner, regulated by agreements not only between 
temporally averaged models but also between cross-architectural models. 
% The testing results outperform the SOTA. 

\section{Experiments}
\subsection{Datasets, metrics, and experimental settings}

\textbf{Datasets.} The DTSL is evaluated on four benchmark datasets. 
These include two 2D datasets, ACDC~\cite{bernard2018deep} and PROMISE12~\cite{litjens2014evaluation}, 
and two 3D datasets, LA~\cite{XIONG2021101832} and Pancreas-NIH~\cite{10.1007/978-3-319-24553-9_68}. 

% The DTSL is evaluated on four 2D and 3D datasets, as they are 2D datasets of
% ACDC~\cite{bernard2018deep}, Promise12~\cite{litjens2014evaluation},
% and 3D datasets of LA~\cite{XIONG2021101832}, Pancreas-NIH~\cite{10.1007/978-3-319-24553-9_68}. 

% Details of the datasets can be found in  Appendix A. 
The \textbf{ACDC} dataset contains multi-slice 2D cine cardiac MRI images from $100$ patients, 
covering four classes: background, right ventricle, left ventricle, and myocardium. 
It is split by patient into $70$ for training, $20$ for validation, and $10$ for testing. 
The \textbf{PROMISE12} dataset, from the MICCAI 2012 challenge, 
contains $50$ prostate MRI images from $50$ patients and is split into $35$ for training, $5$ for validation, and $10$ for testing.
The \textbf{LA} dataset comprises $100$ 3D gadolinium-enhanced MRI (GE-MRI) scans, 
split into $80$ for training and $20$ for testing. 
Finally, the \textbf{Pancreas-NIH} dataset is a 3D contrast-enhanced abdominal CT dataset of $80$ volumes,
which are split into 60 for training and $20$ for testing.

We adopt the data augmentation strategy from BCP~\cite{bai2023bidirectional}. 
This approach defines two distinct levels of transformation: weak augmentation, 
which consists of standard geometric transforms (random rotation and random flips), 
and strong augmentation, which applies color jitter and Gaussian blur in addition to the weak augmentations. 
These two augmentation levels are applied to the inputs of all teacher-student models. 
% In addition, all input data were enhanced using the BCP method. 
% The Pancreas-NIH dataset is a 3D contrast-enhanced abdominal CT dataset containing 
% 80 volumes with manual annotations. 
% We use 60 volumes for training and 20 for testing. 

% Finally, the \textbf{Pancreas-NIH} dataset is a 3D contrast-enhanced abdominal CT dataset of 80 volumes, 
% which are split into 60 for training and 20 for testing.

\textbf{Metrics.} 
Following previous works~\cite{bai2023bidirectional,chi2024adaptive}, 
we adopt four popular evaluation metrics for the ACDC, LA, and Pancreas-NIH datasets: 
\begin{itemize}
    \item the dice similarity coefficient (DSC)~\cite{dice1945measures,sorensen1948method}, 
    which measures the volume overlap, 
    calculated as twice the intersection divided by the sum of the sizes of the predicted and ground truth segmentations; 
    \item the Jaccard index~\cite{jaccard1912distribution}, 
    which quantifies similarity by dividing the intersection by the union of the predicted and ground truth segmentations; 
    \item the 95\% Hausdorff distance (95HD)~\cite{birsan2005one}, 
    which measures the 95th percentile of the maximum distance between the predicted and ground truth surfaces, 
    providing robustness to outliers; 
    \item the average surface distance (ASD)~\cite{celaya2023generalized}, 
    which calculates the average distance between the boundaries of the predicted and ground truth segmentations.
\end{itemize}
For the PROMISE12 dataset, we adhere to its standard evaluation protocol by using DSC and ASD. 
% BCP\cite{bai2023bidirectional} was used as a method for data augmentation.

\textbf{Experimental settings.} 
As aforementioned, the two teacher-student groups employ distinct backbone architectures: 
U-Net~\cite{ronneberger2015u} and ResU-Net~\cite{diakogiannis2020resunet} for the 2D datasets (ACDC, PROMISE12), 
and V-Net~\cite{abdollahi2020vnet} and ResV-Net~\cite{wang2023mcf} for the 3D datasets (LA, Pancreas-NIH).

% For experiments on the 2D datasets, including ACDC and PROMISE12, 
% the U-Net~\cite{ronneberger2015u} and ResU-Net~\cite{diakogiannis2020resunet} are employed 
% as the backbone architectures for the two groups of teacher-student models, respectively. 
% For experiments on the 3D datasets, including LA and Pancreas-NIH, 
% V-Net~\cite{abdollahi2020vnet} and ResV-Net~\cite{wang2023mcf} are used 
% as the backbone architectures for the two groups of models, respectively. 
All experiments were conducted in PyTorch on an NVIDIA RTX 3090 GPU. 
We trained the 2D models (U-Net and ResU-Net) for 30k iterations and the 3D models (V-Net and ResV-Net) for 7.5k iterations. 
The Adam optimizer was used with an initial learning rate of $\eta_0 = 1\times 10^{-3}$, 
which was decayed using a poly learning rate scheduler $\eta=\eta_0\times(1- \mathrm{iter}/\mathrm{max\_iter})^{0.9}$. 
During training, 2D inputs were randomly cropped to $256\times 256$, 
while 3D inputs were cropped to $112\times 112 \times 80$ for the 
LA dataset and $96\times 96 \times 96$ for the Pancreas-NIH dataset.

% Following previous works \cite{bai2023bidirectional,chi2024adaptive},
% we adopt four evaluation metrics for the ACDC, LA, and Pancreas-NIH datasets: 
% Dice Similarity Coefficient (DSC), Jaccard Index, 95\% Hausdorff Distance (95HD), 
% and Average Surface Distance (ASD).
% For the PROMISE12 dataset, we use two metrics: DSC and ASD.

\subsection{Performance gain is primarily attributed to the DTSL framework rather than unlabeled data}
The DTSL framework can be applied
in supervised learning using \Cref{eqn:totallosssupervised} and 
in semi-supervised learning by \Cref{eqn:totallosssupervised,eqn:totallossunsupervised}. 
To disentangle the performance gains from the framework itself versus those from incorporating unlabeled data, 
we conducted a comparative analysis across five experimental settings
\begin{itemize}
\item vanilla supervised learning with limited labeled data; 
\item supervised DTSL with limited labeled data;
\item semi-supervised DTSL with limited labeled data and complementary unlabeled data;
\item vanilla supervised learning with full labeled data; 
\item supervised DTSL with full labeled data. 
\end{itemize}
The results of this analysis are presented in the lower portion of each results 
\Cref{tab:ACDC,tab:Promise12,tab:LA,tab:Pancreas-NIH} 
for the four benchmark datasets. 

% These results indicate that DTSL consistently outperforms the standard U-Net/V-Net baselines. 
These results suggested that, 
when comparing the performance of vanilla supervised learning, supervised DTSL, and semi-supervised DTSL 
under limited labeled data, 
the majority of the performance gain is observed in the transition from vanilla supervised learning to supervised DTSL, 
rather than from supervised DTSL to semi-supervised DTSL. 
This strongly suggests that the performance improvement is primarily driven by the self-paced learning strategy
rather than the incorporation of additional unlabeled data. 

Notably, the semi-supervised DTSL, trained with only limited labeled data, 
surpasses the performance of a standard supervised model trained on the full labeled dataset 
on three of the four benchmarks, namely ACDC, PROMISE12, and Pancreas-NIH. 
This is a interesting finding, as performance on the full dataset is typically considered 
the upper bound for semi-supervised methods. 
This result further validates the effectiveness of our proposed framework. 

% A comparison between the vanilla and DTSL supervised models on limited labeled data reveals a substantial performance gap. 
% This strongly indicates that the primary performance gain is driven by the DTSL framework itself, 
% validating the effectiveness of its self-paced learning mechanism in a limited-data regime. 
% Furthermore, the comparison between supervised DTSL and semi-supervised DTSL shows 
% that incorporating unlabeled samples yields an additional, albeit more modest, 
% performance improvement. 

% To investigate the impact of the self-paced learning mechanism on model performance, 
% we conduct experiments using the DTSL framework in a purely supervised setting across the four datasets, 
% using varying proportions of labeled data, as shown in the bottom parts of \cref{tab:ACDC,tab:Promise12,tab:LA}. 

% These results indicate that DTSL consistently outperforms the standard U-Net/V-Net baselines. 
% Interestingly, when comparing the performance of supervised U-Net/V-Net, supervised DTSL, and semi-supervised DTSL 
% under limited labeled data, 
% the majority of the performance gain is observed in the transition from supervised U-Net/V-Net to supervised DTSL, 
% rather than from supervised DTSL to semi-supervised DTSL. 
% This suggests that the performance improvement is primarily driven by the self-paced learning strategy
% rather than the incorporation of additional unlabeled data.

\subsection{Comparison experiments on benchmark datasets}

\begin{table*}[!htb]
% \caption{Comparison with other methods on the ACDC dataset, \dag  represents supervised training, the total number of training sets is 70, 5\% refers to 3 samples and 10\% refers to 7 samples.}
\caption{Performance comparison with SOTA methods on the ACDC dataset under semi-supervised learning using 5\% and 10\% labeled data. 
Supervised baselines are also included at the bottom for reference.} 
\centering
    {
    \begin{tabular}{rcccccc}
    \toprule
     & \multicolumn{2}{c}{Scans used} & \multicolumn{4}{c}{Metrics} \\
    % \cline{2-7}
    \cmidrule{2-7}
    & Labeled & Unlabeled & DSC$\uparrow$ & Jaccard$\uparrow$ & 95HD$\downarrow$ & ASD$\downarrow$ \\
    \midrule
    DTC (AAAI'21)~\cite{luo2021semi}   & \multirow{8}{*}{3 (5\%)} & \multirow{8}{*}{67 (95\%)} & 56.90 & 45.67 & 23.36 & 7.39 \\
    URPC (MICCAI'21)~\cite{luo2021efficient}    & & & 55.87 & 44.64 & 13.60 & 3.74 \\
    MC-Net (MICCAI'21)~\cite{wu2021semi}    & & & 62.85 & 52.29 & 7.62 & 2.33 \\
    SS-Net (MICCAI'22)~\cite{wu2022exploring}  & & & 65.83 & 55.38 & 6.67 & 2.28 \\
    SCP-Net (MICCAI'23)~\cite{zhang2023self}    & & & 87.27 & - & - & 2.65 \\
    BCP (CVPR'23)~\cite{bai2023bidirectional} & & & 87.59 & 78.67 & 1.90 & 0.67 \\
    ABD (CVPR'24)~\cite{zhao2024alternate} & & & 88.96 & 80.70 & 1.57 & 0.52 \\
    $\beta$-FFT (CVPR'25)~\cite{hu2025beta} & & & 89.58 & 81.68 & \textbf{1.46} & \textbf{0.45}\\
    \textbf{Ours-DTSL} & & & \textbf{90.09} & \textbf{82.43} & 1.74 & 0.54 \\
    \midrule
    DTC (AAAI'21)~\cite{luo2021semi}   & \multirow{10}{*}{7 (10\%)} & \multirow{10}{*}{63 (90\%)} & 84.29 & 73.92 & 12.81 & 4.01 \\
    URPC (MICCAI'21)~\cite{luo2021efficient}  & & & 83.10 & 72.41 & 4.84 & 1.53 \\
    MC-Net (MICCAI'21)~\cite{wu2021semi}   & & & 86.44 & 77.04 & 5.50 & 1.84 \\
    SS-Net (MICCAI'22)~\cite{wu2022exploring}    & & & 86.78 & 77.67 & 6.07 & 1.40 \\
    SCP-Net (MICCAI'23)~\cite{zhang2023self}   & & & 89.69 & - & - & 0.73 \\
    PLGCL (CVPR'23)~\cite{basak2023pseudo}   & & & 89.10 & - & 4.98 & 1.80 \\
    BCP (CVPR'23)~\cite{bai2023bidirectional} & & & 88.84 & 80.62 & 3.98 & 1.17 \\
    ABD (CVPR'24)~\cite{zhao2024alternate} & & & 89.81 & 81.95 & 1.46 & 0.49 \\
    SDCL (MICCAI'24)~\cite{song2024sdcl}  & & & 90.92 & 83.83 & 1.29 & 0.34 \\
    $\mathrm{M^{3}HL}$ (MICCAI'25)~\cite{LiuYaj_M3HL_MICCAI2025} & & & 90.47 & 83.23 & 1.43& 0.34 \\
    $\beta$-FFT (CVPR'25)~\cite{hu2025beta} & & & 90.54 & 83.24 & 1.51 & 0.49\\
    \textbf{Ours-DTSL} & & & \textbf{91.47} & \textbf{84.67} & \textbf{1.10} & \textbf{0.26} \\
    \midrule    
    {U-Net (supervised)}& 3 (5\%) & 0 & 47.83 & 37.01 & 31.16 & 12.62 \\
    {\textsl{Ours-DTSL} (supervised)}& 3 (5\%) & 0 & \textsl{81.36} & \textsl{71.08} & \textsl{6.89} & \textsl{1.56} \\
    \textbf{Ours-DTSL} (semi-supervised)  & 3 (5\%) & 67 (95\%) & \textbf{90.09} & \textbf{82.43} & \textbf{1.74} & \textbf{0.54} \\\midrule
    {U-Net (supervised) }& 7 (10\%) & 0 & 79.41 & 68.11 & 9.35 & 2.70 \\
    {\textsl{Ours-DTSL}} (supervised)& 7 (10\%) & 0 & \textsl{87.81} & \textsl{79.24} & \textsl{4.55} & \textsl{1.23}\\
    \textbf{Ours-DTSL}  (semi-supervised)&7 (10\%) & 63 (90\%)& \textbf{91.47} & \textbf{84.67} & \textbf{1.10} & \textbf{0.26} \\\midrule
    {U-Net (supervised)}& 70 (All) & 0 & 91.44 & 84.59 & 4.30 & 0.99 \\
    {\textsl{Ours-DTSL}} (supervised)& 70 (All) & 0 & \textsl{92.22} & \textsl{85.87} & \textsl{1.71} & \textsl{0.41} \\
\bottomrule
\end{tabular}
}
\label{tab:ACDC}
\end{table*}

\textbf{ACDC dataset.}
We compared our proposed DTSL framework against several SOTA methods on the ACDC dataset, 
including DTC \cite{luo2021semi}, URPC \cite{luo2021efficient}, 
MC-Net \cite{wu2021semi}, SS-Net \cite{wu2022exploring}, SCP-Net \cite{zhang2023self}, 
PLGCL \cite{basak2023pseudo}, BCP \cite{bai2023bidirectional}, ABD \cite{zhao2024alternate}, SDCL \cite{song2024sdcl}, 
$\mathrm{M^{3}HL}$ \cite{LiuYaj_M3HL_MICCAI2025}, and $\beta$-FFT \cite{hu2025beta}.
\Cref{tab:ACDC} presents the average segmentation performance across all classes 
for the 5\%, 10\%, and full labeled data protocols. 
The results demonstrate that DTSL consistently outperforms all SOTA baselines across all four evaluation metrics.

% \cref{tab:ACDC} presents the average segmentation performance across all classes for both the 5\% and 10\% labeled data protocols. 
% The results demonstrate that DTSL consistently outperforms all SOTA baselines across all four evaluation metrics.
% Notably, with only 10\% labeled data, 
% the proposed approach even surpasses fully supervised methods trained on 100\% labeled data 
% in terms of DSC and Jaccard metrics, which is an achievement not matched by any of the compared methods. 

\begin{figure*}[!htb]
    \centering
    \includegraphics[width=0.7\linewidth]{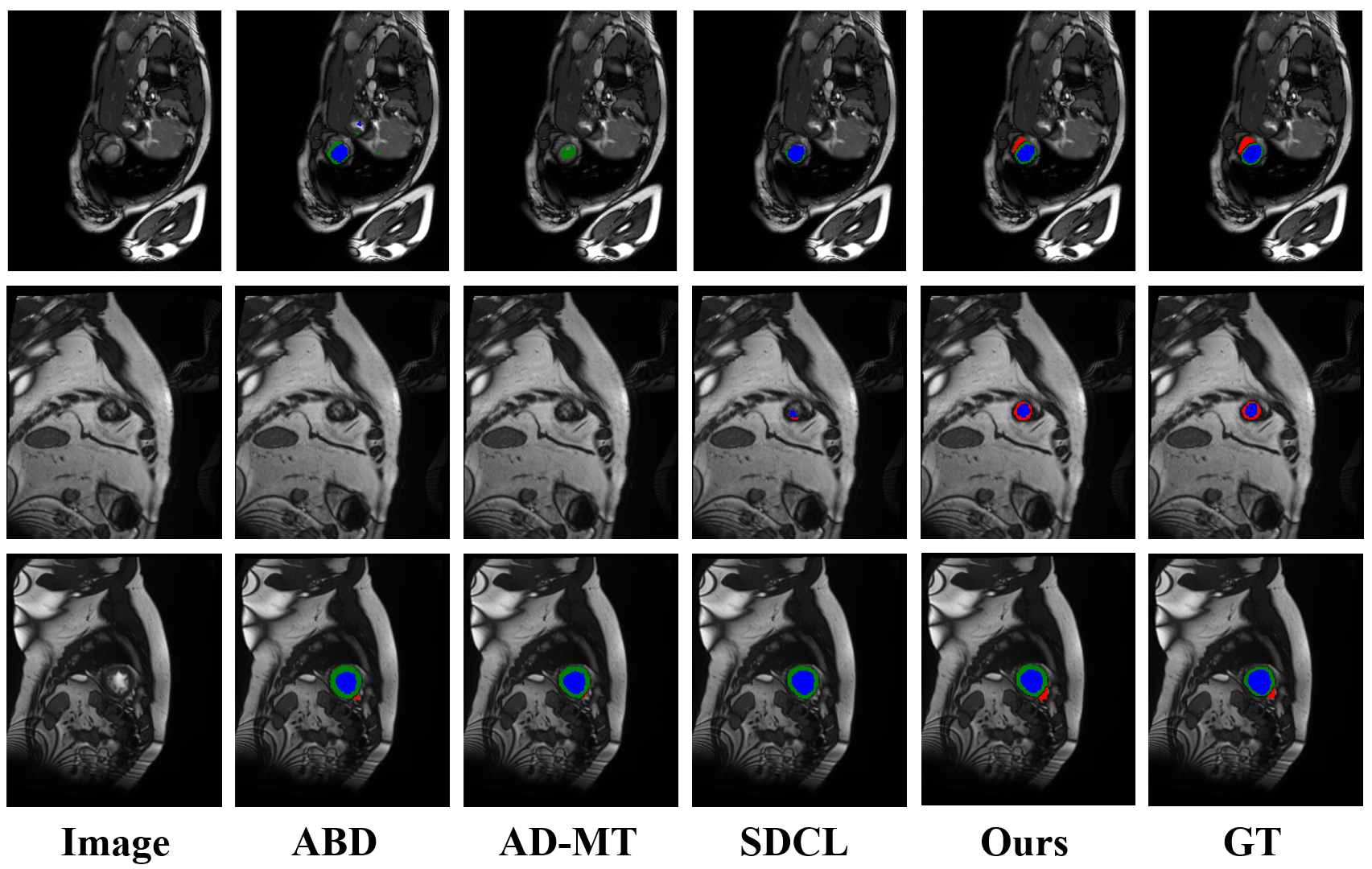}
    \caption{Representative segmentation results on the ACDC dataset using the 10\% labeled data. 
    The proposed method produces segmentations that align more closely with the ground truth than competing approaches, 
    demonstrating superior performance in capturing complex, fine-grained structures. 
    Notably, in the red-highlighted region, only DTSL achieves an accurate segmentation.}
    \label{fig:vis}
\end{figure*}

\Cref{fig:vis} presents representative segmentation results on the ACDC dataset under 
the semi-supervised setting with 10\% labeled data. 
As shown in the figure, 
segmentations produced by the proposed method align more closely with the ground truth
compared to other approaches, 
particularly in capturing fine-grained structures. 
This visual evidence supports the motivation of our self-paced framework, 
which is designed to control the learning pace from easy (e.g., bulk regions) to complex (e.g., fine structures). 
Notably, in the red-highlighted region at the bottom of the figure, only the propsoed method achieves an accurate segmentation.

% demonstrating its ability to minimize prediction errors and produce segmentation results that are closer to the ground truth.

% Notably, with only 10\% labeled data, 
% our approach even surpasses fully supervised methods trained on 100\% labeled data, 
% demonstrating its ability to minimize prediction errors and produce segmentation results that are closer to the ground truth.

% Figure~\ref{fig:vis} shows representative segmentation examples on the ACDC dataset under 
% the semi-supervised setting with 10\% labeled data. As shown in the figure, our method produces 
% segmentation results that are more consistent with the ground truth than other SoTA methods, 
% particularly in fine structures—such as the red region at the bottom of the figure only our method 
% achieves accurate segmentation.

\begin{table*}[!htb]
    \centering
    \caption{Performance comparison with SOTA methods on the PROMISE12 dataset under semi-supervised learning using 
    20\% labeled data. 
Supervised baselines are also included at the bottom for reference.}
    {
    \begin{tabular}{rcccc}
        \toprule
        & \multicolumn{2}{c}{Scans used} & \multicolumn{2}{c}{Metrics} \\
        \cmidrule{2-5}
        & Labeled & Unlabeled & DSC$\uparrow$  & ASD$\downarrow$ \\
        \midrule
        CCT (CVPR'20)~\cite{ouali2020semi}  & \multirow{7}{*}{7 (20\%)} & \multirow{7}{*}{28 (80\%)} & 71.43 & 16.61 \\
        URPC (MICCAI'21)~\cite{luo2021efficient}   & & & 63.23 & 4.33 \\
        SS-Net (MICCAI'22)~\cite{wu2022exploring}   & & & 62.31 & 4.36 \\
        SLC-Net (MICCAI'22)~\cite{liu2022semi}    & & & 68.31 & 4.69  \\
        SCP-Net (MICCAI'23)~\cite{zhang2023self}  & & & 77.06 & 3.52  \\
        ABD (CVPR'24)~\cite{zhao2024alternate}   & & & 82.06 & 1.33 \\
        $\beta$-FFT (CVPR'25)~\cite{hu2025beta}   & & & 84.40 & 1.13 \\       
        \textbf{Ours-DTSL} & & & \textbf{85.19} & \textbf{0.97}  \\
        \midrule
        {U-Net} (supervised)& 7 (20\%) & 0 & 60.88 & 13.87  \\
        {\textsl{Ours-DTSL}} (supervised)& 7 (20\%) & 0 & \textsl{81.61} & \textsl{1.16}  \\
        \textbf{Ours-DTSL} (semi-supervised) & 7 (20\%)&28 (80\%) & \textbf{85.19} & \textbf{0.97}  \\
        \midrule
        {U-Net} (supervised)& 35 (All) & 0 & 84.76 & 1.58  \\
        {\textsl{Ours-DTSL}} (supervised)& 35 (All) & 0 & \textsl{85.73} & \textsl{0.95}  \\
        \bottomrule
    \end{tabular}}
    \label{tab:Promise12}
\end{table*}

\textbf{PROMISE12 dataset.} 
The evaluation results on the PROMISE12 dataset are summarized in \Cref{tab:Promise12}. 
Compared with several SOTA methods, including CCT~\cite{ouali2020semi}, URPC~\cite{luo2021efficient}, 
SS-Net~\cite{wu2022exploring}, SLC-Net~\cite{liu2022semi}, SCP-Net~\cite{zhang2023self}, ABD~\cite{zhao2024alternate}, and $\beta$-FFT \cite{hu2025beta}, 
the proposed DTSL framework demonstrates consistent improvements in both DSC and ASD metrics. 
Notably, in terms of DSC, DTSL achieves a 2.69\% improvement over the second-best method, 
and its ASD is also better than previous approaches.

\begin{table*}[htb]
\centering
\caption{Performance comparison with SOTA methods on the LA dataset under semi-supervised learning using 
    5\% labeled data. 
Supervised baselines are also included at the bottom for reference.}
\label{tab:LA}{
\begin{tabular}{rcccccc}
\toprule
& \multicolumn{2}{c}{Scans used} & \multicolumn{4}{c}{Metrics} \\
\cmidrule{2-7}
& Labeled & Unlabeled & DSC$\uparrow$ & Jaccard$\uparrow$ & 95HD$\downarrow$ & ASD$\downarrow$ \\
\midrule
UA-MT (MICCAI'19)~\cite{yu2019uncertainty} & \multirow{8}{*}{4 (5\%)} & \multirow{8}{*}{76 (95\%)} & 82.26 & 70.98 & 13.71 & 3.82 \\
SASSNet (MICCAI'20)~\cite{li2020shape}  & & & 81.60 & 69.93 & 16.16 & 3.58 \\
DTC (AAAI'21)~\cite{luo2021semi}  & & & 81.25 & 69.33 & 14.90 & 3.99 \\
SS-Net (MICCAI'22)~\cite{wu2022exploring} & & & 86.33 & 76.15 & 9.97 & 2.31 \\
PS-MT (CVPR'22)~\cite{liu2022perturbed}  & & & 88.49 & 79.13 & 8.12 & 2.78 \\
BCP (CVPR'23)~\cite{bai2023bidirectional}  & & & 88.02 & 78.72 & 7.90 &2.15 \\
AD-MT (ECCV'24)~\cite{zhao2024alternate} & & & 89.63 & 81.28 & 6.56 & 1.85 \\
SGRS-Net (MICCAI'25)~\cite{WanTao_SynergyGuided_MICCAI2025} & & & 89.70 & 81.40 & 6.68 & \textbf{1.75} \\
\textbf{Ours-DTSL} & & & \textbf{90.26} & \textbf{82.31} & \textbf{5.34} & 1.94 \\
\midrule
{V-Net (supervised)}& 4 (5\%) & 0 & 52.55 & 39.60& 47.05 & 9.87 \\
{\textsl{Ours-DTSL} (supervised)}& 4 (5\%) & 0 & \textsl{85.71} & \textsl{76.22}& \textsl{8.60} & \textsl{2.72} \\
\textbf{Ours-DTSL} (semi-supervised) & 4 (5\%) &76 (95\%) & \textbf{90.26} & \textbf{82.31} & \textbf{5.34} & \textbf{1.94} \\\midrule
{V-Net (supervised)}& 80 (All) & 0 & 91.47 & 84.36 & 5.48 & 1.51 \\
{\textsl{Ours-DTSL} (supervised)}& 80 (All) & 0 & \textsl{91.64} & \textsl{84.41} & \textsl{5.32} & \textsl{1.41} \\
\bottomrule
\end{tabular}}
\end{table*}

\textbf{LA dataset.} 
As shown in \Cref{tab:LA}, 
the proposed method is compared against several SOTA baselines, 
including UA-MT~\cite{yu2019uncertainty}, SASSNet~\cite{li2020shape}, DTC~\cite{luo2021semi}, SS-Net~\cite{wu2022exploring}, 
PS-MT~\cite{liu2022perturbed}, BCP~\cite{bai2023bidirectional}, 
AD-MT~\cite{zhao2024alternate} and SGRS-Net \cite{WanTao_SynergyGuided_MICCAI2025}. 
The proposed method achieves superior performance on the DSC, Jaccard Index, and 95\% Hausdorff Distance (95HD) metrics. 
While its ASD performance is slightly inferior to that of AD-MT~\cite{zhao2024alternate},
the strong overall results demonstrate the effectiveness and robustness of our approach.

\begin{table*}[htb]
\centering
\caption{Performance comparison with SOTA methods on the Pancreas-NIH dataset under semi-supervised learning using 
    10\% labeled data. 
Supervised baselines are also included at the bottom for reference.}
\label{tab:Pancreas-NIH}{
\begin{tabular}{rcccccc}
\toprule
& \multicolumn{2}{c}{Scans used} & \multicolumn{4}{c}{Metrics} \\
\cmidrule{2-7}
& Labeled & Unlabeled & DSC$\uparrow$ & Jaccard$\uparrow$ & 95HD$\downarrow$ & ASD$\downarrow$ \\
\midrule
UA-MT (MICCAI'19)~\cite{yu2019uncertainty} & \multirow{9}{*}{6 (10\%)} & \multirow{9}{*}{54 (90\%)} & 66.34 & 53.21 & 17.21 & 4.57 \\
SASSNet (MICCAI'20)~\cite{li2020shape}  & & & 68.78 & 53.86 & 19.02 & 6.26 \\
DTC (AAAI'21)~\cite{luo2021semi}  & & & 69.21 & 54.06 & 17.21 & 5.95 \\
ASE-Net (TMI'22)~\cite{lei2022semi}  & & & 71.54 & 56.82 & 16.33 & 5.73 \\
SS-Net (MICCAI'22)~\cite{wu2022exploring} & & & 71.76 & 57.05 & 17.56 & 5.77 \\
PS-MT (CVPR'22)~\cite{liu2022perturbed}  & & & 76.94 & 62.73 & 13.12 & 3.66 \\
BCP (CVPR'23)~\cite{bai2023bidirectional} & & & 73.83 & 59.24 & 12.71 & 3.72 \\
AD-MT (ECCV'24)~\cite{zhao2024alternate} & & & 80.21 & 67.51 & 7.18 & 1.66 \\
SGRS-Net (MICCAI'25)~\cite{WanTao_SynergyGuided_MICCAI2025}   & & & 80.55 & 67.88 & 6.00 & 2.50\\
\textbf{Ours-DTSL} & & & \textbf{83.07} & \textbf{71.24} & \textbf{4.79} & \textbf{1.30}\\
\midrule
{V-Net (supervised)}& 6 (10\%) & 0 & 55.60 & 41.74 & 45.33 & 18.63 \\
{\textsl{Ours-DTSL} (supervised)}& 6 (10\%) & 0 & \textsl{73.32} & \textsl{63.42} & \textsl{8.59} & \textsl{5.13} \\
\textbf{Ours-DTSL} (semi-supervised) & 6 (10\%) & 54 (90\%) & \textbf{83.07} & \textbf{71.24} & \textbf{4.79} & \textbf{1.30}\\
\midrule
{V-Net (supervised)}& 60 (All) & 0 & 82.60 & 70.81 & 5.61 & 1.33 \\
{\textsl{Ours-DTSL} (supervised)}& 60 (All) & 0 & \textsl{83.51} & \textsl{71.62} & \textsl{4.71} & \textsl{1.25} \\
\bottomrule
\end{tabular}}
\end{table*}

\textbf{Pancreas-NIH dataset.} 
\Cref{tab:Pancreas-NIH} presents the results on the Pancreas-NIH dataset, 
comparing the proposed method against several SOTA baselines, including 
UA-MT~\cite{yu2019uncertainty},
SASSNet~\cite{li2020shape}, DTC~\cite{luo2021semi}, ASE-Net~\cite{lei2022semi},SS-Net~\cite{wu2022exploring},
PS-MT~\cite{liu2022perturbed}, BCP~\cite{bai2023bidirectional},AD-MT~\cite{zhao2024alternate}, 
SGRS-Net\cite{WanTao_SynergyGuided_MICCAI2025}.
The results demonstrate that DTSL outperforms all these methods across all evaluated metrics.

\subsection{Ablation studies and hyperparameter optimization}
We conducted ablation studies on the ACDC dataset using the 10\% labeled data, 
unless otherwise stated, 
to evaluate the effectiveness of each proposed DTSL module, different CLG strategies, 
and the impact of hyperparameter optimization.
% More comprehensive ablation studies can be found in Appendix B. 

\textbf{Effectiveness of each module in DTSL.} 
% Experiments are conducted on a vanilla MT strategy, 
% a dual pseudo-label setup using teacher and across-group student models (plain DTSL without CLG), 
% the use of CLG for pseudo-label generation as described in \cref{eqn:CLG}, 
% and the application of URL in \cref{eqn:URL}, 
% with the results presented in \cref{tab:ablation}. 
As presented in \Cref{tab:ablation}, the ablation study compares the performance of several configurations: 
a vanilla MT strategy; a `plain DTSL' setup (dual pseudo-labels from teacher and across-group student models without CLG); 
DTSL enhanced with the CLG module in \Cref{eqn:CLG}; and the full framework including the URL in \Cref{eqn:URL}.

The results in \Cref{tab:ablation} indicate that employing dual teacher-student groups (Plain DTSL) significantly 
outperforms the standard single-group MT approach. 
We attribute this improvement to the implicit self-paced learning introduced by using dual pseudo-labels. 
Similar to our analysis of the supervised MT strategy in \Cref{eqn:mt1,eqn:mt2,eqn:mt3}, 
consensus between the pseudo-labels reinforces the learning signal, 
while disagreement reflects competing gradients that blur the optimization target. 

Furthermore, incorporating the CLG module, which acts as a precise pace-regulator, 
substantially enhances performance over the Plain DTSL. 
Finally, applying the URL on top of CLG provides an additional, though smaller, improvement.

\begin{table*}[!htb]
    \centering
    \caption{
        Ablation study on the effectiveness of each DTSL component using the ACDC 10\% dataset. 
        The table compares the performance of: (1) a vanilla MT baseline, (2) a `Plain DTSL' setup (dual pseudo-labels without CLG/URL), 
        (3) DTSL enhanced with the CLG module, and (4) the full DTSL framework including both CLG and URL. 
    % Effectiveness of of each module in DTSL.
    }
    % \belowrulesep=0pt
    % \aboverulesep=0pt
    {%\footnotesize
    \begin{tabular}{cccc|cccc}
    \toprule
    MT & Plain DTSL & CLG & $\mathcal{L}_{\mathrm{URL}}$  & DSC$\uparrow$ & Jaccard$\uparrow$ & 95HD$\downarrow$ & ASD$\downarrow$\\
       \midrule
       \checkmark &&&&80.10& 75.60& 4.31& 2.33\\
       &\checkmark  &  &  & 89.73 & 81.97 & 1.71 &  0.49\\
       &     & \checkmark &  & 91.26 & 84.34 & 1.40 &  0.43\\
       &\checkmark  &  & \checkmark & 88.65 & 80.44 & 2.26 &  0.67\\
       &     & \checkmark & \checkmark & \textbf{91.47} & \textbf{84.67} & \textbf{1.10} & \textbf{0.26}\\
       \bottomrule
    \end{tabular}
    }
    
    \label{tab:ablation}
\end{table*}

\textbf{Results on different CLG strategies.} 
To validate the choice of inputs for the CLG, we explored several alternative strategies. 
Taking $\mathrm{student}_0$ as an example, 
the available candidate signals are its in-group teacher ($\mathrm{teacher}_0$), 
the cross-group student ($\mathrm{student}_1$), 
and the cross-group teacher ($\mathrm{teacher}_1$). 
We compared the following setups (a symmetric configuration is used for $\mathrm{student}_1$), 
\begin{itemize}
    \item Strategy 1: consensus between $\mathrm{student}_1$ and $\mathrm{teacher}_1$; 
    \item Strategy 2: consensus between $\mathrm{teacher}_0$ and $\mathrm{teacher}_1$;
    \item Strategy 3: consensus among $\mathrm{student}_1$, $\mathrm{teacher}_0$, and $\mathrm{teacher}_1$;
    \item Default: consensus between $\mathrm{teacher}_0$ and $\mathrm{student}_1$.
\end{itemize}
As shown in \Cref{tab:CLG}, the default setting achieves the best performance. 
This result validates the motivation to control the learning pace is by leveraging the consensus between the 
temporally-lagged signal and the cross-architectural signal.

% Taking $\mathrm{student}_0$ as an example, 
% the proposed method (default) generates pseudo-labels using outputs from 
% $\mathrm{teacher}_0$ and $\mathrm{student}_1$ via the CLG module. 
% However, there are three different models $\mathrm{student}_1$, $\mathrm{teacher}_0$, and $\mathrm{teacher}_1$ 
% could be candidated as the input of the CLG module. 
% In view of this, alternative CLG strategies, including  
% using $\mathrm{teacher}_1$ and $\mathrm{student}_1$ to generate pseudo-labels 
% for $\mathrm{student}_0$ (Strategy 1), 
% using $\mathrm{teacher}_0$ and $\mathrm{teacher}_1$ to generate pseudo-labels 
% for $\mathrm{student}_0$ (Strategy 2), 
% and using triple outputs of $\mathrm{teacher}_0$, $\mathrm{teacher}_1$, and $\mathrm{student}_1$ to generate pseudo-labels
% for $\mathrm{student}_0$ (Strategy 3), 
% are further explored. 
% The training process for $\mathrm{student}_1$ follows a similar setup. 
% As shown in \cref{tab:CLG}, the default setting achieves the best performance among all strategies, 
% indicating that the engaging the temporally averaged model and the cross-architectural model as the signals that 
% controls the learning pace archieves the best results. 

\begin{table*}[!htb]
    % \centering
    % \noindent
    % \begin{minipage}{\textwidth}
    \caption{
        Ablation study validating the input strategy for the Consensus Label Generator (CLG). 
        The table compares our `Default' setup against alternative signal combinations on the ACDC 10\% dataset.
    % Results of different strategies of CLG.
    }\label{tab:CLG}
    \centering
    {
    \begin{tabular}{rcccc}
    \toprule
       Strategy   & DSC$\uparrow$ & Jaccard$\uparrow$ & 95HD$\downarrow$ & ASD$\downarrow$\\
       \midrule
       Strategy 1 & 90.21 & 82.69 & 1.70 & 0.53\\
       Strategy 2  & 90.45 & 83.21 & 1.98 & 0.41\\
       Strategy 3 & 90.64 & 83.29 & 1.60 & 0.30\\
       \textbf{Default} & \textbf{91.47} & \textbf{84.67} & \textbf{1.10} & \textbf{0.26}  \\
       \bottomrule
    \end{tabular}
    }
\end{table*}    

% Table~\ref{tab:CLG} shows the effectiveness of CLG's different strategies. 
% Among them, 'Normal' refers to training Student2 model using pseudo labels generated by Student1 model 
% and Teacher1 model without using the CLG module, similarly, training Student1 model using pseudo labels 
% generated by Student2 model and Teacher2 model, below we will only introduce the label generation process used 
% for training Student2 model, ignoring Student1 model.  'Strategy 1' means that we use CLG to get the pseudo labels 
% generated by Student1 model and Teacher1 model into a label that can be used for Student2 model training. 
% 'Strategy 2' is that we use all three labels generated by Student1 model, Teacher1 model and Teacher2 model 
% to get a new label to train Studen2 model.

\textbf{Results on alternative network architectures.} 
To explore the impact of architectural choice on DTSL, 
we evaluated several combinations for the two model group, 
U-Net/ResU-Net (default), U-Net/Swin-Unet~\cite{cao2022swin}, and ResU-Net/Swin-Unet. 
The comparison, conducted on the ACDC 10\% labeled dataset, is presented in \Cref{tab:structure}.
The results confirm that the default U-Net/ResU-Net configuration achieves the best performance among the tested combinations.

\begin{table*}[!htb]
\centering
\caption{Impact of network architecture combinations on performance, evaluated on the ACDC 10\% dataset. 
The default (U-Net/ResU-Net) is compared to U-Net/Swin-Unet and ResU-Net/Swin-Unet.
    % The results of various architecture combinations on 10\% labeled samples from the ACDC dataset.
    }
{
\begin{tabular}{rcccccc}
\toprule
Architectures & DSC$\uparrow$ & Jaccard$\uparrow$ & 95HD$\downarrow$ & ASD$\downarrow$\\
\midrule
U-Net/SwinU-net  & 88.67 & 80.73 & 2.81 & 0.92 \\
ResU-Net/SwinU-net  &  89.09 & 81.13 & 2.67 & 0.77 \\
\textbf{U-Net/ResU-net}  &  \textbf{91.47} & \textbf{84.67} & \textbf{1.10} & \textbf{0.26}\\
\bottomrule
\end{tabular}}
\label{tab:structure}
\end{table*}

\subsection{Hyperparameter optimization}

\textbf{Optimization of the threshold $\kappa$.} 
The threshold $\kappa$, introduced in \Cref{eqn:kappa}, is a vital hyperparameter as it determines the boundary 
between consistent (easy) and inconsistent (difficult) regions based on JS divergence, 
which are critical for regulating the self-paced learning curriculum. 
Experimental results, presented in \Cref{tab:kappa}, 
show that performance generally peaks at a specific value; it increases as $\kappa$ rises from 0.05 and 
then decreases at higher values. 
We found that the optimal value for $\kappa$ is consistently around $0.1$ across all four datasets.

% And $\kappa$ on other datasets are shown in \cref{tab:lambda-acdc5,tab:lambda-la,tab:lambda-promise,tab:lambda-pancreas}.

% As shown in \cref{tab:lambda-acdc5,tab:lambda-la,tab:lambda-pancreas,tab:lambda-promise}, 
% the optimal value of the hyperparameter $\kappa$ consistently lies around 0.1 
% across all four datasets when using the commonly used proportion of labeled samples. 

\begin{table*}[htb]
    \centering
    \caption{Hyperparameter optimization for the JS divergence threshold $\kappa$ in \Cref{eqn:kappa}, 
    evaluated on the ACDC 10\% dataset. 
    Optimal performance is achieved at $\kappa = 0.1$ across all four datasets.}
    {\setlength{\tabcolsep}{1.2pt}\footnotesize
    \begin{tabular}{r|cccc|cccc|cccc|cccc}
        \toprule
        \multicolumn{1}{c}{} & \multicolumn{4}{c}{ACDC (5\% labeled)}                                  & \multicolumn{4}{c}{PROMISE12 (20\% labeled)}                             & \multicolumn{4}{c}{LA (5\% labeled)}                                   & \multicolumn{4}{c}{Pancreas-NIH (10\% labeled)}                         \\
        % \cmidrule{2-5}\cmidrule{5-9}
        $\kappa$             & DSC$\uparrow$  & Jaccard$\uparrow$ & 95HD$\downarrow$ & ASD$\downarrow$ & DSC$\uparrow$  & Jaccard$\uparrow$ & 95HD$\downarrow$ & ASD$\downarrow$ & DSC$\uparrow$  & Jaccard$\uparrow$ & 95HD$\downarrow$ & ASD$\downarrow$ & DSC$\uparrow$  & Jaccard$\uparrow$ & 95HD$\downarrow$ & ASD$\downarrow$ \\
        \midrule
        0.05                 & 89.85          & 82.11             & \textbf{1.35}    & \textbf{0.36}   & 82.57          & 70.44             & 3.41             & 1.34            & 89.55          & 81.16             & 8.91             & 3.03            & 82.33          & 70.20             & 5.19             & 1.34            \\
        0.10                 & \textbf{90.09} & \textbf{82.43}    & 1.74             & 0.54            & 84.28          & 72.98             & \textbf{2.54}    & \textbf{0.94}   & \textbf{90.26} & \textbf{82.31}    & \textbf{5.34}    & \textbf{1.94}   & \textbf{83.07} & \textbf{71.24}    & \textbf{4.69}    & \textbf{1.30}   \\
        0.15                 & 89.32          & 81.35             & 2.13             & 0.62            & \textbf{85.19} & \textbf{74.35}    & 2.60             & 0.97            & 89.77          & 81.50             & 5.61             & 2.21            & 82.19          & 69.98             & 5.33             & 1.52            \\
        0.20                 & 89.58          & 81.76             & 2.12             & 0.58            & 83.81          & 72.33             & 3.75             & 1.40            & 89.56          & 81.20             & 6.42             & 2.38            & 83.05          & 71.21             & 4.94             & 1.32            \\
        \bottomrule
    \end{tabular}
    }
\label{tab:kappa}
\end{table*}

\begin{table*}[htb!]
% \begin{minipage}{\textwidth}
% \begin{minipage}[t]{0.47\textwidth}
\caption{Hyperparameter optimization for the $\mathcal{L}_{\mathrm{semi}}$ weight $\alpha$ in \Cref{eqn:pace}, 
conducted with $\beta$ fixed at $0.05$, evaluated on the ACDC 10\% dataset. 
Optimal performance is achieved around $\alpha=1.00$ for 2D datasets (ACDC, PROMISE12) 
and $\alpha=0.10$ for 3D datasets (LA, Pancreas-NIH).
% Optimal performance is consistently achieved around $\alpha=1.00$ 
% for 2D datasets (ACDC, PROMISE12) and $\alpha=0.10$ for 3D datasets (LA, Pancreas-NIH).
% Optimization of $\alpha$ on 10\% labeled ACDC dataset.
}
\centering
\setlength{\tabcolsep}{1.2pt}
{\footnotesize
\begin{tabular}{r|cccc|r|cccc|r|cccc|r|cccc}
\toprule
\multicolumn{5}{c}{ACDC (10\% labled)}                                                  & \multicolumn{5}{c}{PROMISE12 (20\% labeled)}                                            & \multicolumn{5}{c}{LA (5\%)}                                                            & \multicolumn{5}{c}{Pancreas-NIH (10\% labeled)}                                         \\

$\alpha$      & DSC$\uparrow$  & Jac.$\uparrow$ & 95HD$\downarrow$ & ASD$\downarrow$ & $\alpha$      & DSC$\uparrow$  & Jac.$\uparrow$ & 95HD$\downarrow$ & ASD$\downarrow$ & $\alpha$      & DSC$\uparrow$  & Jac.$\uparrow$ & 95HD$\downarrow$ & ASD$\downarrow$ & $\alpha$      & DSC$\uparrow$  & Jac.$\uparrow$ & 95HD$\downarrow$ & ASD$\downarrow$ \\\midrule
0.50          & 91.18          & 84.22             & 1.94             & 0.45            & 0.50          & 84.08          & 72.73             & 4.08             & 1.38            & 0.05          & 90.05          & 81.99             & 7.15             & 2.51            & 0.05          & 82.52          & 70.43             & 4.90             & 1.26            \\
0.75          & 91.31          & 84.40             & 1.13             & 0.40            & 0.75          & 84.76          & 73.70             & 3.09             & 1.17            & 0.10          & 90.26          & 82.31             & 5.34             & 1.94            & \textbf{0.10} & \textbf{83.07} & \textbf{71.24}    & 4.79             & 1.30            \\
\textbf{1.00} & \textbf{91.47} & \textbf{84.67}    & \textbf{1.10}    & \textbf{0.26}   & 1.00          & 84.69          & 73.63             & 2.64             & 1.05            & 0.15          & 89.61          & 81.29             & 6.42             & 2.05            & 0.15          & 83.02          & 71.17             & 4.70             & 1.25            \\
1.25          & 90.71          & 83.56             & 1.13             & 0.32            & \textbf{1.25} & \textbf{85.19} & \textbf{74.35}    & \textbf{2.60}    & \textbf{0.97}   & \textbf{0.20} & \textbf{89.80} & \textbf{81.60}    & \textbf{6.14}    & \textbf{2.20}   & 0.20          & 82.80          & 70.87             & 4.87             & 1.26            \\
1.50          & 90.11          & 82.66             & 1.22             & 0.29            & 1.50          & 84.42          & 73.21             & 2.66             & 1.00            & 0.30          & 90.02          & 81.93             & 6.20             & 2.20            & 0.30          & 82.95          & 71.10             & \textbf{4.65}    & \textbf{1.24}   \\
\bottomrule
\end{tabular}
}
\label{tab:alpha}
\end{table*}

\textbf{Optimization of $\alpha$ in \Cref{eqn:pace}.}
The pace regulator, $\mathcal{L}_{\mathrm{pace}}$, is composed of $\mathcal{L}_{\mathrm{semi}}$ and $\mathcal{L}_{\mathrm{URL}}$, 
weighted by hyperparameters $\alpha$ and $\beta$, respectively. 
Since the full supervised loss $\mathcal{L}^{\ell}$ also includes $\mathcal{L}_{\mathrm{sup}}$, 
$\alpha$ and $\beta$ should be tuned as independent hyperparameters.

However, considering the unlabeled loss $\mathcal{L}^{u}$ consists solely of $\mathcal{L}_{\mathrm{pace}}$ and 
dominates the overall training objective in SSMIS, 
we found the relative ratio between $\alpha$ and $\beta$ to be more critical than their absolute values. 
To provide a clear analysis, we adopted a standard tuning strategy by fixing one hyperparameter while varying the other.

As shown in \Cref{tab:alpha}, with $\beta=0.05$ fixed, 
the results demonstrate that the optimal choice for $\alpha$ is around $1.00$ for the 2D datasets (ACDC, PROMISE12) 
and $0.10$ for the 3D datasets (LA, Pancreas-NIH).

\textbf{Optimization of $\omega$ of the EMA process.} 
In the MT framework, the weights of the teacher model are updated via the EMA
of the student model's weights as defined in \Cref{eqn:EMA},  
making the teacher a temporally smoothed version of the student. 
The update factor $\omega$ controls this temporal smoothing, 
and we hypothesized it might be critical for regulating the learning pace. 

To investigate this effect, we conducted experiments on the ACDC 10\% dataset, 
varying $\omega$ from $0.90$ to $0.99$. 
The results, presented in \Cref{tab:omega}, 
show that while optimal performance was achieved at $\omega=0.95$, 
the overall performance variation across the tested range is minimal. 
This suggests that the DTSL framework is robust to the specific choice of $\omega$. 

\begin{table*}[htb!]
    \centering
    \caption{
        Hyperparameter optimization for the EMA update factor $\omega$ in \Cref{eqn:EMA}, 
        evaluated on the ACDC 10\% dataset. 
        Although optimal performance is observed at $\omega = 0.95$, 
        the minimal performance variation demonstrates that the DTSL framework is highly robust to this parameter.}
    {
    \begin{tabular}{rccccccccccc}
        \toprule
        $\omega$ & 0.90 & 0.91  & 0.92  & 0.93  & 0.94 & \textbf{0.95} & 0.96  & 0.97  & 0.98  & 0.99  \\\midrule
        DSC      & 91.20  & 91.12 & 91.19 & 91.17 & 91.15  & \textbf{91.33} & 91.30 & 91.29 & 91.24 & 91.26 \\
        \bottomrule
    \end{tabular}
    }
    \label{tab:omega}
\end{table*}

\section{Conclusion}
Inspired by interpreting the MT strategy by self-paced learning, 
we propose the DTSL framework for SSMIS. 
The framework employs two groups of teacher-student models with distinct architectures. 
A CLG is introduced to produce pseudo-labels by leveraging both outputs from temporally averaged and cross-architectural models. 
These pseudo-labels guide the learning process from easy to complex regions. 
Additionally, a URL is applied to uncertain regions, 
encouraging the model to reconsider low-confidence predictions and further refine the learning pace. 
Experimental results demonstrated
that our method achieves SOTA performance on four benchmark datasets. 
Remarkably, on three of the four datasets, 
our semi-supervised method with limited labeled data surpasses its fully supervised counterparts. 

%%%%%%%%%%%%%%%%%%%%%%%%%%%%%%%%%%%%%%%%%%%%%%%%%%%%%%%
%%% Supplements. 补充材料, 非必选
%%%%%%%%%%%%%%%%%%%%%%%%%%%%%%%%%%%%%%%%%%%%%%%%%%%%%%%
% \Supplements{Appendix A.}

%%%%%%%%%%%%%%%%%%%%%%%%%%%%%%%%%%%%%%%%%%%%%%%%%%%%%%%
%%% Open Access Funding Note. 开放获取基金来源说明
%%%%%%%%%%%%%%%%%%%%%%%%%%%%%%%%%%%%%%%%%%%%%%%%%%%%%%%
%\FundingNote{\textbf{Open Access} funding enabled and organized by CAUL and its Member Institutions.}

% \clearpage
%%%%%%%%%%%%%%%%%%%%%%%%%%%%%%%%%%%%%%%%%%%%%%%%%%%%%%%
%%% Reference section. 参考文献
%%% Citation in the content using "some words~\cite{1,2}".
%%% ~ is needed to make the reference number is on the same line with the word before it.
%%% Using scis.bst to format the style if the ref.bib file is included, e.g.,
% \bibliographystyle{scis}
\bibliographystyle{unsrt}
\bibliography{ref}
%%%%%%%%%%%%%%%%%%%%%%%%%%%%%%%%%%%%%%%%%%%%%%%%%%%%%%%
% \begin{thebibliography}{99}

% \bibitem{1} Author A, Author B, Author C. Reference title. Journal, 2024, 38: 13--28

% \bibitem{2} Author A, Author B, Author C, et al. Reference title. In: Proceedings of Conference, Place, 2024. 6--12

% \end{thebibliography}

%%%%%%%%%%%%%%%%%%%%%%%%%%%%%%%%%%%%%%%%%%%%%%%%%%%%%%%
%%% Appendix sections. 附录章节, 非必选
%%%%%%%%%%%%%%%%%%%%%%%%%%%%%%%%%%%%%%%%%%%%%%%%%%%%%%%
%\begin{appendix}
%\section{Name}

%\end{appendix}

\end{document}